\pdfoutput=1

\documentclass[11pt,table,xcdraw]{article}

\usepackage{acl}

\usepackage{times}
\usepackage{latexsym}
\usepackage{multirow}
\usepackage{booktabs}
\usepackage[T1]{fontenc}

\usepackage[utf8]{inputenc}
\usepackage{amsfonts}
\usepackage{amsmath}
\usepackage{amssymb}
\usepackage{microtype}

\usepackage{inconsolata}

\usepackage{graphicx}

\usepackage{amssymb}
\usepackage{graphicx}
\usepackage{marvosym}
\usepackage{url}
\usepackage{fancyhdr}
\usepackage{hyperref}
\usepackage{lipsum}
\usepackage{enumitem}
\usepackage{booktabs}
\usepackage{algorithm}
\usepackage{algorithmic}
\usepackage{amsmath}
\usepackage{subcaption}
\usepackage{pifont}
\usepackage{colortbl}   

\usepackage[most]{tcolorbox}

\tcbset{
  parskip/.style={
    before={\par\pagebreak[0]\parindent=0pt},
    after={\parfillskip=0pt\par},
  },
}
\newtcolorbox{resultbox}[1][]{%
    colback=gray!30,
    colframe=black!3,
    notitle,
    sharp corners,
    borderline west={2pt}{0pt}{gray!80!black},
    enhanced,
    breakable,
    boxsep=0pt,
    left=4pt,right=2pt,top=2pt,bottom=2pt,
    }

\definecolor{class5}{RGB}{255, 204, 204} 
\definecolor{class4}{RGB}{255, 255, 204} 
\definecolor{class3}{RGB}{204, 255, 204} 
\definecolor{class2}{RGB}{204, 255, 255} 
\definecolor{class1}{RGB}{204, 204, 255} 
\definecolor{class0}{RGB}{255, 204, 255} 
\definecolor{NavyBlue}{RGB}{40, 160, 250} 

\newcommand{\sbs}[1]{%
    \pgfmathsetmacro{\normalized}{int((#1 - 0.4) / (1.0 - 0.4) * 100)}%
    \edef\tempa{\noexpand\cellcolor{NavyBlue!\normalized}}%
    \tempa#1%
}

\newcommand{\sbsalt}[1]{%
    \pgfmathsetmacro{\normalized}{int(min(100, max(0, #1 * 100)))}%
    \edef\tempa{\noexpand\cellcolor{NavyBlue!\normalized}}%
    \tempa#1%
}

\newcommand{\redx}{\cellcolor{white!40!}\ding{55}} 
\newcommand{\greentick}{\cellcolor{white!90!}\ding{51}} 

\usepackage[normalem]{ulem}

\title{\texttt{TigerLLM} - A Family of Bangla Large Language Models}

\author{Nishat Raihan \\
  George Mason University \\
  Fairfax, VA, USA \\
  \texttt{mraihan2@gmu.edu} \\\And
  Marcos Zampieri \\
  George Mason University \\
  Fairfax, VA, USA \\
  \texttt{mzampier@gmu.edu} \\}

\begin{document}
\maketitle
\begin{abstract}
The development of Large Language Models (LLMs) remains heavily skewed towards English and a few other high-resource languages. This linguistic disparity is particularly evident for Bangla - the $5^{th}$ most spoken language. A few initiatives attempted to create open-source Bangla LLMs with performance still behind high-resource languages and limited reproducibility. To address this gap, we introduce \texttt{TigerLLM} - a family of Bangla LLMs. Our results demonstrate that these models surpass all open-source alternatives and also outperform larger proprietary models like GPT3.5 across standard benchmarks, establishing \texttt{TigerLLM} as the new baseline for future Bangla language modeling.
\end{abstract}

\section{Introduction}

LLMs have fundamentally transformed NLP by achieving exceptional performance across a broad range of tasks \cite{brown2020language, chowdhury2022palm, raihan2025large}. While these models exhibit unprecedented capabilities in language understanding, generation, reasoning, and specialized applications, their advancements predominantly benefit high-resource languages \cite{alam2024llms}. This inequality is particularly noticeable for Bangla. Despite having about 237 million native speakers,\footnote{\url{ethnologue.com/language/ben/}} Bangla remains quite underserved in modern NLP advancements.

This under-representation stems primarily from the limitation of high-quality training data. While proprietary models like GPT-4 \cite{openai2023gpt4} and Claude-3.5 \cite{anthropic2024claude35} demonstrate reasonable Bangla capabilities, open-source alternatives consistently underperform. Recent multilingual models such as Gemma-2 \cite{team2024gemma} and LLaMA 3.1 \cite{dubey2024llama3}, despite leveraging diverse training corpora and advanced tokenization systems like TikTokenizer \cite{corso2024we}, also fail to deliver satisfactory performance for Bangla.

\begin{table*}[!t]
\centering
\scalebox{0.73}{
    \begin{tabular}{l c c c c c c c c c}
    \toprule
     & \textbf{Base-LLM} & \textbf{Size} & \textbf{pt} & \textbf{corpora} & \textbf{ft}  & \textbf{ft-dataset} & \textbf{Paper/Report?} & \textbf{Reproducibility?} \\ 
    \midrule
    \href{https://huggingface.co/hishab/titulm-gemma-2-2b-v1.1}{titu-Gemma} & Gemma-2 & 2B & 4.4B & \redx & \redx & \redx & \redx & \redx \\
    \href{https://huggingface.co/hishab/titulm-llama-3.2-3b-v1.1}{titu-LLaMA} & LLaMA-3.1 & 3B & 37B  & \redx & \redx & \redx & \redx& \redx \\
    \href{https://huggingface.co/BanglaLLM/BanglaLLama-3.2-3b-bangla-alpaca-orca-instruct-v0.0.1}{Bangla-LLaMA} & LLaMA-3.2 & 3B & \greentick  & \redx & 172K & Orca-translated & \greentick& \redx \\
    \href{https://huggingface.co/KillerShoaib/gemma-2-9b-bangla-4bit}{G2B} & Gemma-2 & 9B & \redx  & \redx & 145K & Alpaca-translated & \redx& \redx \\
    \href{https://huggingface.co/BanglaLLM/bangla-llama-13b-instruct-v0.1}{Bangla-LLaMA} & LLaMA-2 & 13B & \greentick & \redx & 145K & Alpaca-translated & \redx & \redx \\
    \midrule
    \texttt{TigerLLM} & LLaMA-3.2 & 1B & 10M & \texttt{Bangla-TextBook} & 100K & \texttt{Bangla-Instruct} & \greentick & \greentick \\
    \texttt{TigerLLM} & Gemma-2 & 9B & 10M & \texttt{Bangla-TextBook} & 100K & \texttt{Bangla-Instruct} & \greentick & \greentick \\
    \bottomrule
    \end{tabular}
}
\caption{Comparative analysis of Bangla LLM initiatives and their methodological approaches. The pretraining (\textit{pt}) and finetuning (\textit{ft}) columns indicate corpus size in tokens and instruction count respectively. 
}
\label{table_models}
\end{table*}

\subsection{Limitations of Bangla LLM Initiatives}
\label{sec:limitations}


\paragraph{Training} Recent attempts at developing Bangla LLMs (see Table \ref{table_models}) through continual pretraining (\href{https://huggingface.co/hishab/titulm-gemma-2-2b-v1.1}{titu-Gemma}) and model distillation approaches \cite{zehady2024bongllama} have yielded low and non-reproducible results (see Table \ref{tabl1_results}), often performing worse than their base models. The absence of technical documentation and academic publications further compounds this issue by making result reproduction impossible. Our investigation into these models' performances reveals the need for improvement in the training process. While the unavailability of pretraining corpora limits our analysis of that phase, the finetuning approach demonstrates consistent problematic patterns. 

\paragraph{Data} Most Bangla LLM initiatives rely on translated versions of synthetic datasets like Alpaca-Instruct \cite{taori2023alpaca} and OpenOrca \cite{mitra2023orca}, which are generated through model distillation \cite{hinton2015distilling}. This approach suffers from two fundamental limitations: (1) the datasets are generated by early GPT-3.5 \cite{brown2020language} releases, a model with limited Bangla support, resulting in suboptimal instruction quality, and (2) these English datasets are translated to Bangla using machine translation systems like \href{https://translate.google.com/}{Google Translate} with limited quality checks, further degrading the training data quality. These cascading compromises in training data ultimately result in poor model performance.

\subsection{Contributions}

To address the recurring challenges in Bangla LLM development, we introduce three fundamental contributions:

\begin{enumerate}[nosep,leftmargin=*]
    \item The {\bf \texttt{Bangla-TextBook}} corpus, comprising 10 million tokens of carefully curated educational content across multiple domains, prioritizing content quality over scale.
    \item A high-quality {\bf \texttt{Bangla-Instruct}} dataset of 100 thousand instruction-response pairs, generated through self-instruct \cite{selfinstruct} and model distillation using state-of-the-art teacher models (GPT-4o and Claude-3.5-Sonnet).
    \item The {\bf \texttt{Tiger-LLM}} family (1B and 9B parameters), featuring models pretrained and finetuned on our high-quality datasets, achieving 30-55\% performance improvements over existing benchmarks.
\end{enumerate}

\noindent All components are open-sourced to establish robust foundations for future Bangla language modeling research.\footnote{\url{https://github.com/mraihan-gmu/TigerLLM/tree/main/}}

\begin{figure*}[!t]
    \centering
    \includegraphics[width=0.95\linewidth]{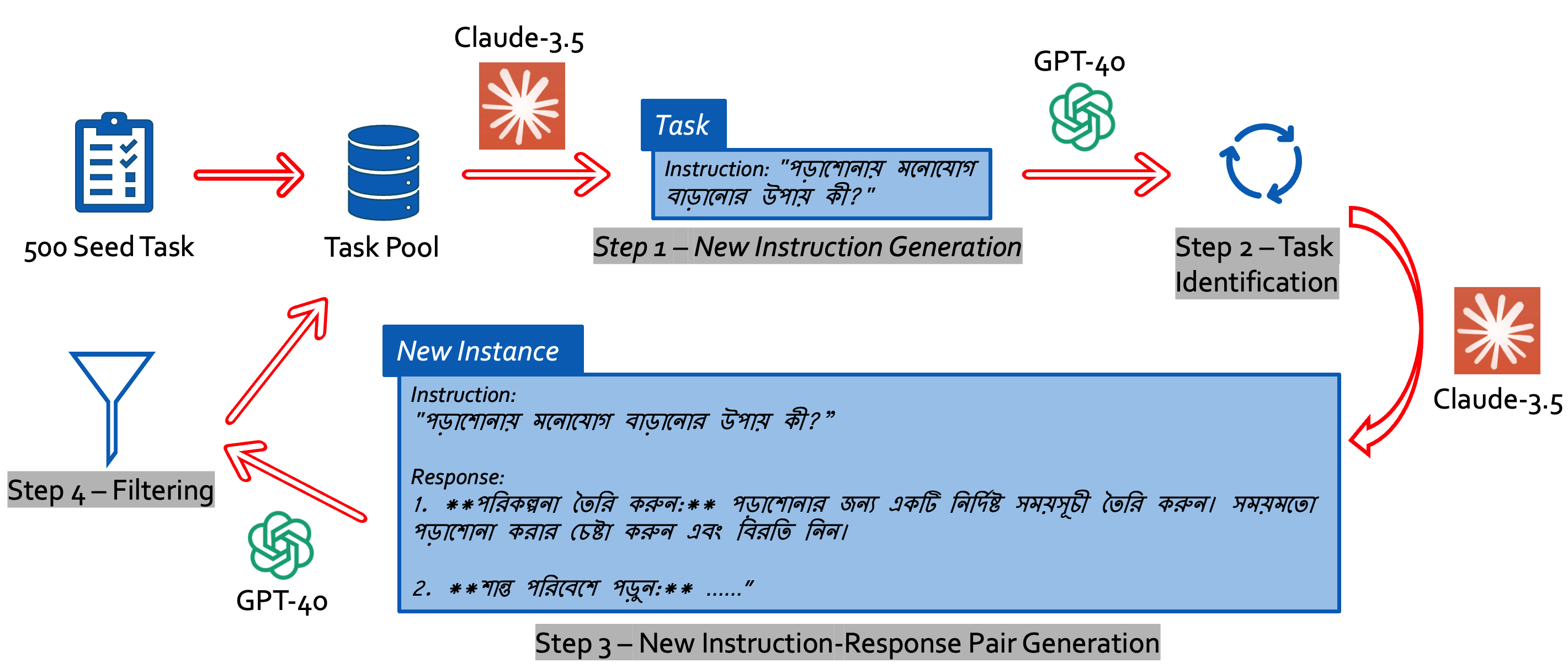}
    \caption{The Bangla-Instruct generation pipeline. With 500 seed tasks, we employ a multi-step process using GPT-4o and Claude-3.5-Sonnet as teacher models to generate instruction-response pairs in Bangla.}
    \label{fig:bangla-instruct}
\end{figure*}

\section{Related Work}

Early transformer-based \textit{encoder-only} pre-trained language models such as BERT \cite{devlin2019bert} concentrate on high‑resource languages like English. Subsequent work adapts them to mid‑ and low‑resource contexts through continued pre‑training and task‑specific finetuning. In Bangla, for instance, \citet{kowsher2022bangla} present \textsc{BanglaBERT}, demonstrating that a dedicated monolingual encoder markedly improves downstream classification and QA relative to multilingual baselines. 

The shift to \textit{decoder‑only} models has produced large multilingual models — e.g.\ \textsc{BLOOM} \cite{le_scao2022bloom}, \textsc{Llama3} \cite{dubey2024llama3}, and \textsc{Aya} \cite{ustun2024aya}—that cover dozens of under‑represented languages. Yet empirical analyses reveal that these models still perform best when prompted in high‑resource languages, with significant degradation for languages such as Bangla or Swahili \cite{raihan-etal-2025-mhumaneval,jin2024better}. 



As discussed in the previous section, dedicated Bangla decoder models remain scarce and fragmented. GPT2‑Bangla \cite{bhattacharjee2023banglagpt} continues GPT‑2 pre‑training on a 4GB Bangla corpus, while Bong‑\textsc{LLaMA} \cite{zehady2024bongllama} and the \textit{titu‑Gemma}\footnote{\url{https://huggingface.co/hishab/titulm-gemma-2-2b-v1.1}} checkpoint attempt instruction tuning on translated datasets. These efforts often lack rigorous evaluation protocols, transparent data curation, or reproducible training pipelines—as reflected in the inconsistent results summarized in Table~\ref{table_models}. Consequently, a clear methodological gap persists in developing open, reproducible decoder‑only LLMs that natively support Bangla and other low‑resource languages.

\section{\texttt{Bangla-TextBook} Corpus}

Previous Bangla LLMs rely predominantly on corpora sourced from OSCAR \cite{ortiz-suarez-etal-2020-monolingual} and \href{https://commoncrawl.org/}{Common Crawl} \cite{bhattacharjee-etal-2022-banglabert, zehady2024bongllama}, despite quality control challenges. While alternative Bangla corpora have emerged \cite{bhattacharya2023vacaspati}, the absence of curated educational content remains a critical gap. This emphasis on data quality is particularly significant given recent findings by \citet{gunasekar2023textbooks} and \citet{raihan-etal-2025-mojobench}, which demonstrate that LLMs achieve superior performance through high-quality training data, even with reduced volume.

\noindent To bridge this gap, we present the \texttt{Bangla-TextBook} corpus, constructed exclusively from high-quality \textbf{open-source} educational materials published by the \href{https://nctb.gov.bd/}{National Curriculum and Textbook Board} of Bangladesh. We collect texts from 163 textbooks for Grades 6-12, resulting in a total of 9,897,623 tokens and 697,903 sentences. 


\section{\texttt{Bangla-Instruct}} \label{instruct}

To address the limitations described in Section \ref{sec:limitations}, we introduce Bangla-Instruct, a collection of 100,000 native Bangla instruction-response pairs bootstrapped using self-instruct \cite{selfinstruct}. While instruction datasets like Alpaca \cite{taori2023alpaca} and OpenOrca \cite{mitra2023orca} utilized GPT3 and GPT3.5 respectively, we significantly improve upon their approach by employing GPT-4 and Claude-3.5-Sonnet as our teacher models, leveraging their superior instruction-following capabilities.

Our dataset creation begins with 500 diverse seed tasks carefully curated by a team of 50 undergraduate and graduate students from leading Bangladeshi universities (Appendix \ref{app:volunteer}). These volunteers, spanning various academic disciplines and geographical regions of Bangladesh, ensure our seed tasks capture authentic linguistic patterns and cultural contexts. Each seed task undergoes multiple rounds of peer review to maintain quality and cultural sensitivity. Further information on quality control is presented in Appendix (Appendix \ref{app:filter}).

Our generation pipeline consists of four primary steps, each designed to maintain data quality and cultural authenticity (see Figure \ref{fig:bangla-instruct}). 

\noindent \textbf{(1) Seed \& Instruction Generation:}  We begin with a human‑curated seed pool $\mathcal{T}_{s}=\{t_1,\dots,t_{500}\}$ drawn from \emph{50} volunteers representing five academic disciplines across Bangladesh (see Appendix~\ref{app:volunteer}). At every generation round~$i$, we sample $k=8$ seed tasks and prompt \textsc{Claude} to create a candidate batch of instructions $\mathcal{I}_n$, expanding coverage of the ten seed categories $c_{1\dots10}$ listed in Appendix~\ref{app:seed} while preserving authentic linguistic patterns.

\noindent \textbf{(2) Task Typing:}  
    Each instruction $i\in\mathcal{I}_n$ is classified by GPT‑4o into $\tau(i)\in\{\textit{open‑ended},\textit{classification},\textit{generation}\}$, providing the expected answer style and the minimum‑length threshold $l_{\min}(\tau)$ used in subsequent filtering.

\noindent \textbf{(3) Response Drafting:}  
    Conditioned on $(i,\tau(i))$, \textsc{Claude} produces a comprehensive response $r_i$.  
    We retain the highest‑scoring draft according to an internal coherence metric $c(i,r)$.

\noindent \textbf{(4) Multi‑stage Filtering:}  
    GPT‑4o applies the four‑criteria filter $\mathcal{F}$—Language~($\mathcal{L}$), Cultural~($\mathcal{C}$), Quality~($\mathcal{Q}$), and Novelty~($\mathcal{N}$) (see Appendix~\ref{app:filter}).  
    On average, \textasciitilde63\% of $(i,r)$ pairs pass $\mathcal{F}$, yielding a balanced complexity mix (40\% basic, 40\% intermediate, 20\% advanced).  Valid pairs are appended to $\mathcal{T}_{s}$, and the loop continues until 100K high‑quality instruction–response pairs are reached.

By coupling two complementary LLMs with strict verification and a human‑seeded, domain‑balanced task pool, our pipeline mitigates error propagation and preserves cultural nuance—addressing shortcomings observed in earlier Bengali instruction datasets (see Appendix~\ref{app:instruct} for full statistics).

\section{\texttt{TigerLLM}}
As candidate base models, we consider 3 families of multilingual LLMs - LLaMA 3.2 (1B, 3B) \cite{dubey2024llama3}, Gemma-2 (2B, 9B) \cite{team2024gemma} and Pangea (7B) \cite{yue2024pangea}. 

\begin{figure}[!h]
    \centering
    \includegraphics[width=0.8\linewidth]{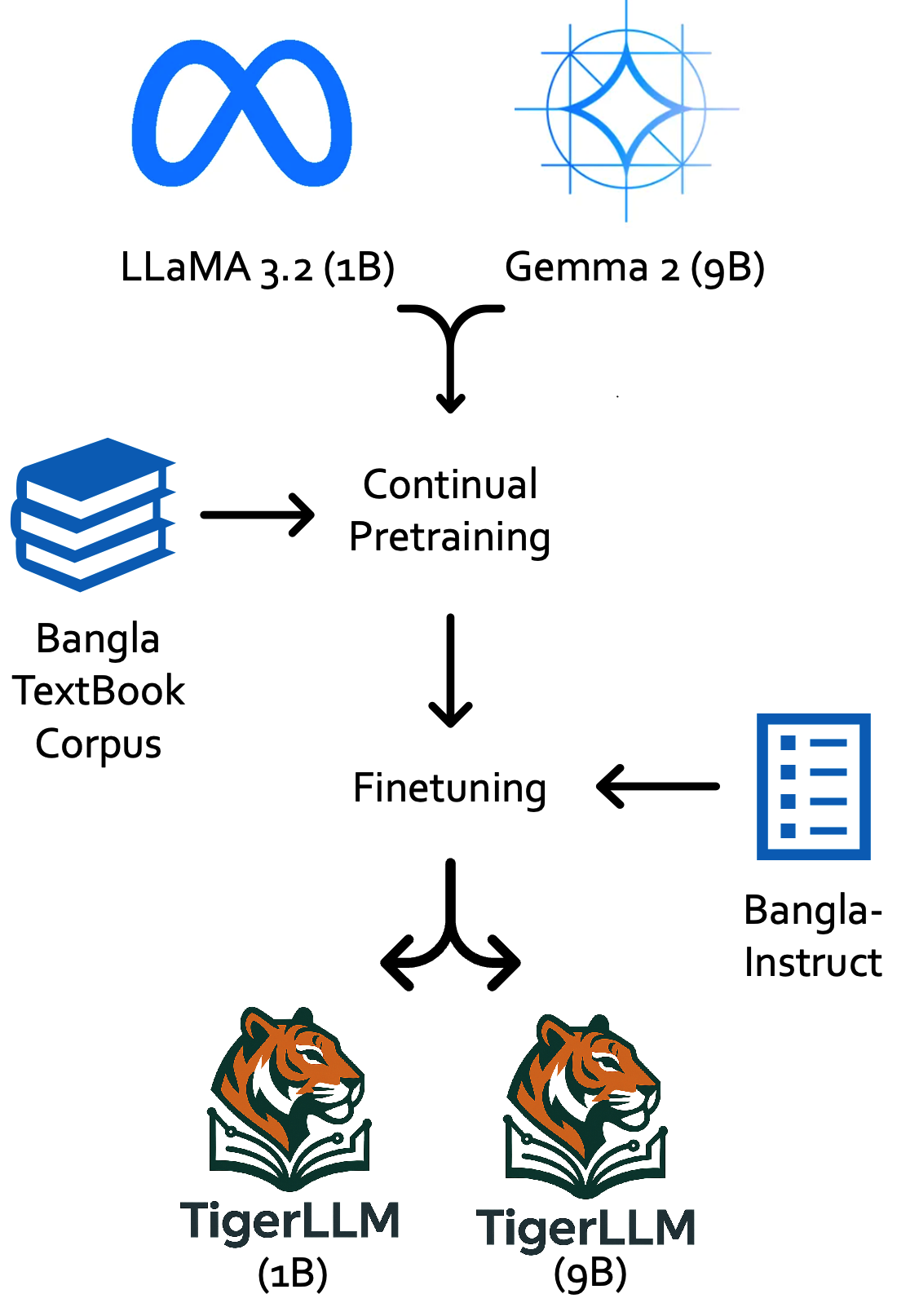}
    \caption{Evolution of TigerLLM.}
    \label{fig:model}
\end{figure}

\paragraph{Evolution of TigerLLM} Figure \ref{fig:model} depicts the final selection of the models and a high-level overview of the process.

\begin{figure}[!h]
    \centering
    \includegraphics[width=0.9\linewidth]{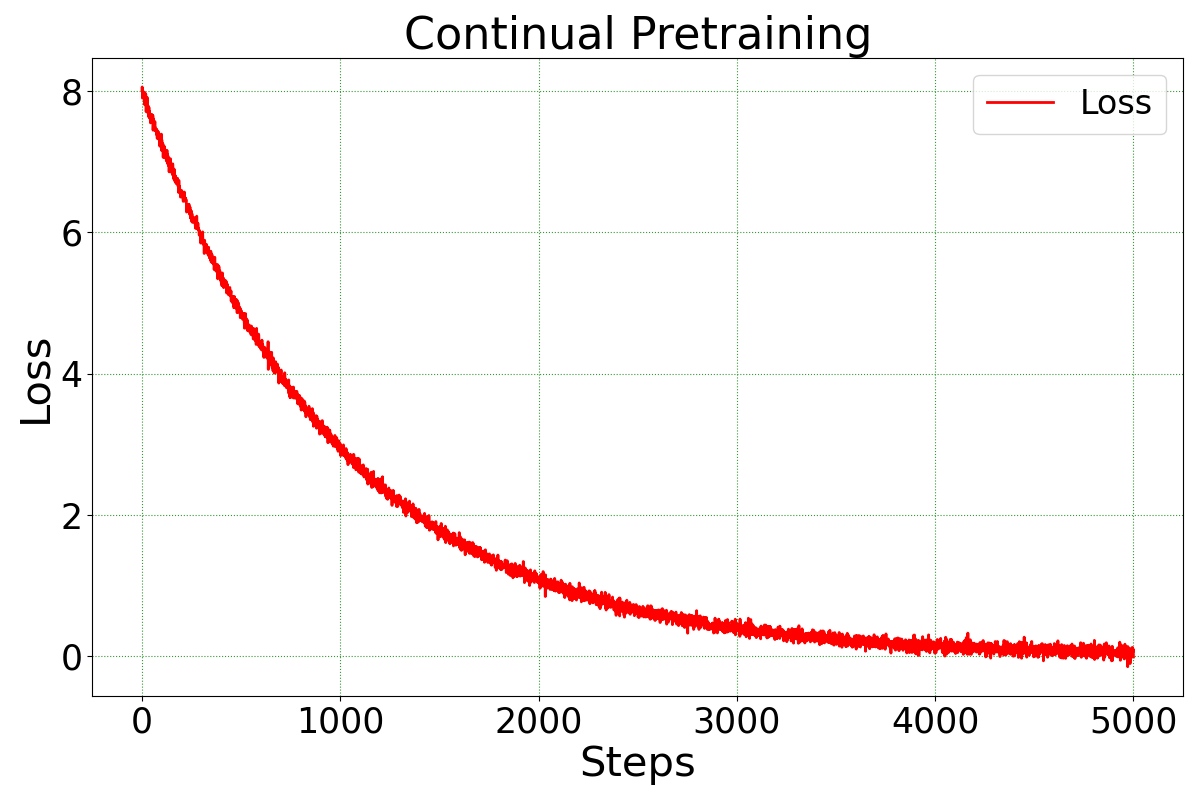}
    \caption{Continual Pretraining - Loss per Steps.}
    \label{fig:cpt}
\end{figure}

\noindent Upon the selection phase, we finalize two pre-trained language models—LLaMA 3.2 (1B) and Gemma 2 (9B)—chosen for their robust foundational capacities. These models then undergo continual pretraining (see Figure \ref{fig:cpt}) on a specialized \textbf{Bangla-TextBook} corpus, which infuses them with a richer understanding of the Bangla language, including its context-specific nuances, stylistic variations, and domain-specific terminology.

\paragraph{Pretraining} We utilize a computing cluster with 8 NVIDIA A100 GPUs (40GB each), 512GB RAM, and 2TB storage. The distributed training setup enables efficient parallel processing, completing the pretraining in approximately 120 hours on this high-performance configuration with gradient checkpointing enabled.

\paragraph{Continual Pretraining} We use the \texttt{Bangla-TextBook} corpus for the models to learn culture and language-specific nuances and gather sufficient and reliable knowledge from a set of high-quality texts. The pretraining phase has been carried out multiple times with empirical choices of hyper-parameters. 

\paragraph{Finetuning} We conduct finetuning on a single NVIDIA A100 (40GB) through Google Colab\footnote{\url{colab.research.google.com}}, supported by 80GB RAM and 256GB storage. The process completes in approximately 96 hours, proving sufficient for model adaptation and task-specific optimization with minimal computational overhead.

\begin{figure}[!h]
    \centering
    \includegraphics[width=0.9\linewidth]{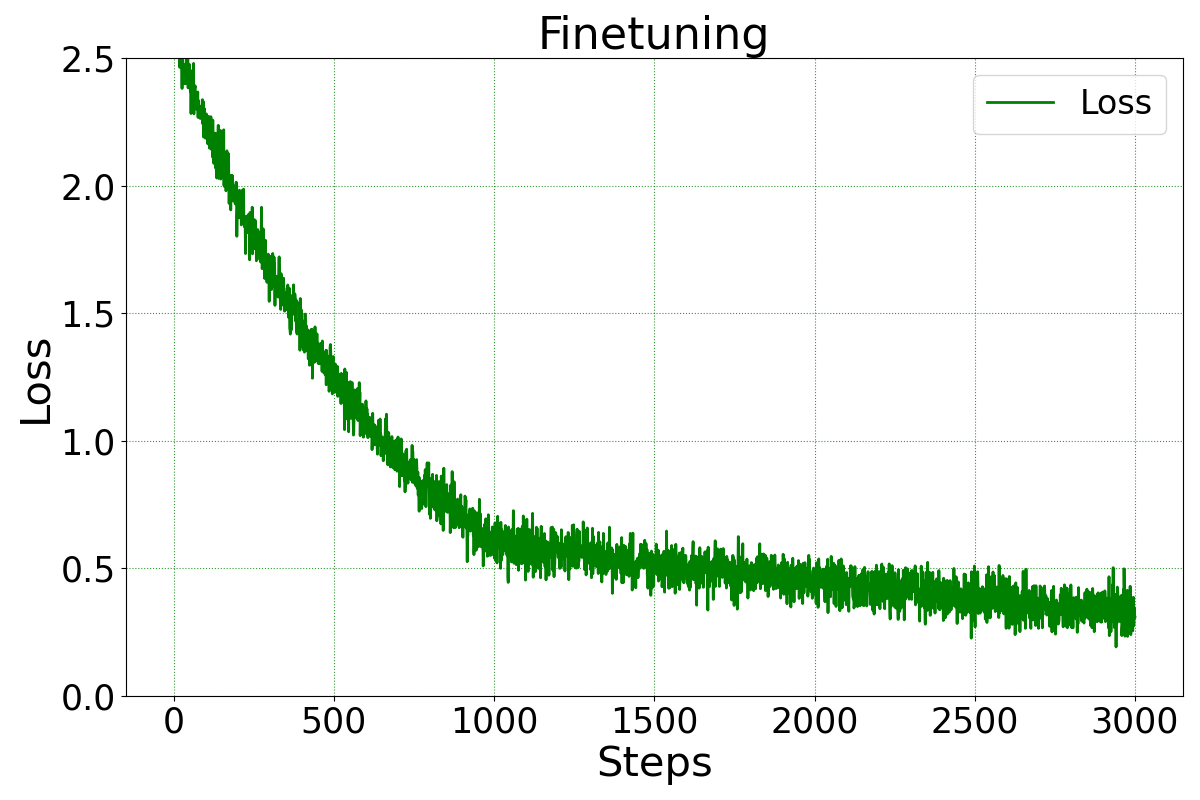}
    \caption{Finetuning - Loss per Steps.}
    \label{fig:ft}
\end{figure}

\paragraph{Model Distillation}  Following this continual pretraining step, the models are finetuned on a carefully curated \textbf{Bangla-Instruct} dataset (Figure \ref{fig:ft}). LoRA \cite{hu2021lora} is not used, we implement full finetuning for better learning. To speed up the training process, we utilize Flash Attention \cite{dao2022flashattention}, we set key parameters: 2048 token maximum sequence length, batch size of 8, 4 gradient accumulation steps, and 3 epochs. Learning rate ($5 \times 10^{-5}$), weight decay (0.02), and 10\% warm-up steps ensure stable convergence. Table \ref{tab:hyperparameters2} in Appendix \ref{app:params} lists complete hyperparameters. 

\begin{table*}[!t]
\centering
\scalebox{0.8}{
    \begin{tabular}{l c c c c c c}
    \toprule
     & \textbf{MMLU-bn} & \textbf{PangBench-bn} & \textbf{BanglaQuaD} & \textbf{mHumanEval-bn}  & \textbf{BEnQA} & \textbf{BanglaRQA} \\ 
    \midrule
    & \textit{understanding} & \textit{multitasking}  & \textit{question answering} & \textit{coding} & \textit{knowledge} & \textit{reasoning} \\ 
    \midrule
    \midrule
    GPT3.5 & 0.55 & 0.55 & 0.50 & 0.56 & 0.50 & 0.49 \\ 
    Gemini-Flash1.5 & 0.66 & 0.57 & 0.62 & 0.58 & 0.56 & 0.61 \\ 
    GPT4o-mini & 0.67 & 0.62 & 0.65 & 0.56 & 0.60 & 0.60 \\ 
    \midrule
    LLaMA3.2 (11B) & 0.22 & 0.19 & 0.21 & 0.15 & 0.18 & 0.20 \\ 
    Gemma 2 (27B) & 0.35 & 0.51 & 0.43 & \textbf{0.64} & 0.50 & 0.56 \\ 
    Pangea (7B) & 0.18 & 0.15 & 0.17 & 0.10 & 0.14 & 0.16 \\ 
    \midrule
    Titu-LLM & 0.06 & 0.19 & 0.08 & 0.02 & 0.17 & 0.21 \\ 
    Bong-LLaMA & 0.05 & 0.12 & 0.08 & 0.02 & 0.15 & 0.13 \\ 
    Bangla-LLaMA & 0.02 & 0.08 & 0.05 & 0.10 & 0.11 & 0.09 \\ 
    Bangla-Gemma & 0.18 & 0.15 & 0.12 & 0.10 & 0.22 & 0.19 \\ 
    \midrule
    \midrule
    TigerLLM (1B) & 0.61 & 0.55 & 0.68 & 0.61 & 0.59 & 0.62 \\ 
    TigerLLM (9B) & \textbf{0.72} & \textbf{0.68} & \textbf{0.70} & 0.63 & \textbf{0.65} & \textbf{0.68} \\ 
    \bottomrule
    \end{tabular}
}
\caption{Performance comparison of \texttt{TigerLLM} with other models on various Bangla-specific benchmarks. All values are reported as \textbf{\% in Pass@1}, where higher scores indicate better performance.}
\label{tabl1_results}
\end{table*}

\noindent By blending the foundational strengths of LLaMA and Gemma with specialized Bangla corpora and instruction-oriented finetuning, the final TigerLLM models emerge as optimized solutions capable of delivering high-quality, instruction-following responses tailored to Bangla-language tasks.

\section{Evaluation}

\paragraph{Bangla LLM Benchmarks} Although there has been limited research on Bangla LLMs, several benchmarks have been established to assess their performance. We focus on five benchmarks specifically curated to evaluate Bangla LLMs across a diverse set tasks. For multitask understanding, we use the Bangla subset of MMLU-Pro \cite{wang2024mmlu} and PangBench \cite{yue2024pangea}. For question answering, we consider BanglaQuaD \cite{Rony2024BanglaQuADAB}, while for general knowledge, we use BEnQA \cite{shafayat2024benqa}. For reasoning tasks, we refer to BanglaRQA \cite{ekram2022banglarqa}. 

As shown in the survey of \citet{raihan2024code}, most coding benchmarks like HumanEval \cite{chen2021evaluating} do not support Bangla, so we utilize the Bangla subset of mHumanEval \cite{raihan-etal-2025-mhumaneval}. 

\paragraph{Results} We present the results obtained by the two TigerLLM models compared to a variety of strong LLM baselines in Table \ref{tabl1_results}. The performance comparison of various models on Bangla-specific benchmarks reveals a common trend. The fine-tuned models generally perform worse than their base counterparts across most tasks. In particular, the results reported by the authors are not reproducible, as mentioned in Section \ref{sec:limitations}. However, \texttt{TigerLLM} is the only finetuned model, consistently outperforming both its base and fine-tuned variants across all tasks. Even the 1B variant does better than most models, falling short to only its 9B counterpart, further validating our emphasis on high-quality data (Section \ref{instruct}).

\paragraph{Takeaways}
TigerLLM demonstrates that carefully curated, high-quality datasets can yield superior performance even with smaller model sizes. Our results show that the 1B parameter model outperforms larger alternatives across multiple benchmarks, emphasizing the importance of data quality over quantity. The success of our Bangla-TextBook corpus and Bangla-Instruct dataset establishes a new paradigm for low-resource language model development.

\section{Conclusion and Future Work}

This paper introduces TigerLLM, a family of state-of-the-art Bangla language models that outperforms existing alternatives across six benchmarks. TigerLLM's success stems from two key innovations: (1) the high-quality Bangla-TextBook corpus derived from educational materials and (2) the carefully curated Bangla-Instruct dataset generated using advanced teacher models. 

The three resources introduced here (corpus, instruction dataset, and models) establish a robust foundation for future Bangla language modeling research. Together, they will contribute to speeding up advances in Bangla language modeling. 

In future work we will conduct a deeper qualitative analysis of the model’s behavior, broaden the corpus to cover a wider array of domains, scale the model to larger parameter counts without compromising quality, and devise richer evaluation metrics tailored specifically to Bangla tasks.

\section*{Limitations}

While TigerLLM delivers state-of-the-art performance, several limitations warrant acknowledgment. First, our Bangla-TextBook corpus, though carefully curated, is limited to educational materials from grades 6-12, potentially missing broader linguistic patterns present in other domains. The 10 million token size, while sufficient for our current models, may constrain scaling to larger architectures. Additionally, our Bangla-Instruct dataset, despite its quality-focused generation process, covers only a subset of possible instruction types and may not fully capture the complexity of real-world Bangla language use cases.

Furthermore, our models are currently limited to 1B and 9B parameters, primarily due to computational constraints and our emphasis on thorough experimentation with smaller computationally efficient architectures. While this approach enabled rapid iteration and quality-focused development, it may not fully exploit the potential benefits of larger model scales. 

\section*{Ethical Considerations}
\label{sec:ethics}

Our work prioritizes ethical considerations throughout the development process. The Bangla-TextBook corpus uses open-source publicly available educational materials from the National Curriculum and Textbook Board of Bangladesh. The volunteer-driven seed task creation process incorporated diverse perspectives while maintaining cultural sensitivity and avoiding harmful biases.

We implemented rigorous filtering mechanisms to ensure cultural appropriateness, gender neutrality, and religious sensitivity in our instruction dataset. The multi-stage review process, involving both automated checks and human verification, helps prevent the propagation of harmful stereotypes or biases. Additionally, our open-source approach promotes transparency and enables community oversight of model behavior.

We strongly recommend that users implement appropriate safeguards when deploying TigerLLM in production environments, particularly for applications involving sensitive information or critical decision-making. 

\bibliography{custom}

\clearpage
 
\appendix

\section{\texttt{Bangla-Instruct} Curation}
\label{app:instruct}

\subsection{Volunteer Information}
\label{app:volunteer}

The seed tasks were created by 50 undergraduate and graduate students from various universities across Bangladesh, ensuring geographical and academic diversity:
\begin{itemize}[nosep]
    \item 15 students from Computer Science and Engineering.
    \item 10 students from Bengali Literature.
    \item 10 students from Business Administration.
    \item 8 students from Science and Engineering.
    \item 7 students from Social Sciences.
\end{itemize}

Each volunteer contributed 10 diverse instructions, resulting in our initial pool of 500 seed tasks. The distribution ensured coverage across multiple domains while preserving authentic Bengali linguistic patterns and cultural contexts.

\subsection{The Seed Dataset}
\label{app:seed}

Our seed dataset comprises 10 distinct categories, carefully chosen to cover a broad spectrum of tasks relevant to Bengali language and culture:

\begin{enumerate}
    \item \textbf{Cultural Knowledge and Heritage} ($c_1$): Tasks focusing on Bengali traditions, festivals, folk tales, and historical events. These include explaining cultural practices, describing traditional ceremonies, and discussing historical significance of various customs.
    
    \item \textbf{Academic Writing} ($c_2$): Structured writing tasks ranging from essay outlines to full academic compositions. Topics cover various academic disciplines while maintaining Bengali writing conventions and scholarly standards.
    
    \item \textbf{Mathematical Problem Solving} ($c_3$): Tasks involving mathematical concepts explained in Bengali, including algebra, geometry, and arithmetic. Special attention is given to Bengali mathematical terminology and local problem-solving contexts.
    
    \item \textbf{Programming and Technical} ($c_4$): Programming problems described in Bengali with solutions in standard programming languages. Includes algorithm explanation, code documentation, and technical concept elaboration in Bengali.
    
    \item \textbf{Creative Writing} ($c_5$): Open-ended creative tasks including story writing, poetry composition, and descriptive passages. Emphasizes Bengali literary devices, metaphors, and cultural storytelling elements.
    
    \item \textbf{Scientific Explanation} ($c_6$): Tasks requiring clear explanation of scientific concepts in Bengali, focusing on making complex ideas accessible while maintaining technical accuracy. Covers physics, chemistry, biology, and environmental science.
    
    \item \textbf{Business and Economics} ($c_7$): Professional writing tasks including business case analyses, market reports, and economic concept explanations. Incorporates local business contexts and Bengali business terminology.
    
    \item \textbf{Social Issues Analysis} ($c_8$): Critical analysis tasks addressing contemporary social issues in Bangladesh and Bengali society. Includes problem identification, cause analysis, and solution proposition.
    
    \item \textbf{Data Analysis and Statistics} ($c_9$): Tasks involving interpretation and analysis of data presented in Bengali, including statistical concepts explanation, data visualization description, and numerical analysis.
    
    \item \textbf{Language and Translation} ($c_{10}$): Tasks focused on Bengali language mastery, including idiom explanation, translation between Bengali and English, and linguistic analysis of Bengali texts.
\end{enumerate}

Each category accounts for approximately 10\% of the seed dataset ($50 \pm 5$ tasks per category), ensuring balanced representation across domains. The tasks within each category vary in complexity level: 40\% basic, 40\% intermediate, and 20\% advanced, based on linguistic complexity and cognitive demand.

\subsection{Filtering Methodology}
\label{app:filter}

Our filtering process $\mathcal{F}: (\mathcal{I}, \mathcal{R}) \rightarrow \{0,1\}$ implements the following criteria:

\begin{enumerate}
    \item \textbf{Language Adherence} ($\mathcal{L}$)
    \begin{itemize}[nosep]
        \item Bengali Word Ratio: $\frac{|\text{Bengali Words}|}{|\text{Total Words}|} \geq 0.95$
        \item Unicode Consistency: $\forall c \in \text{text}, c \in \text{Bengali-UTF8}$
        \item Grammar Check: Using GPT-4o's Bengali grammar scoring function $g(x) \geq 0.8$
    \end{itemize}
    
    \item \textbf{Cultural Sensitivity} ($\mathcal{C}$)
    \begin{itemize}[nosep]
        \item Religious Neutrality: $r(x) \in [-0.1, 0.1]$ on our bias scale
        \item Regional Inclusivity: No specific region/dialect preference
        \item Gender Representation: Balanced pronouns and roles
        \item Political Neutrality: Avoidance of partisan content
    \end{itemize}
    
    \item \textbf{Content Quality} ($\mathcal{Q}$)
    \begin{itemize}[nosep]
        \item Minimum Length: $l(x) \geq l_{min}(\tau)$ where $\tau$ is task type
        \item Coherence Score: $c(i,r) \geq 0.8$ between instruction $i$ and response $r$
        \item Factual Accuracy: Verified against Bengali Wikipedia
        \item Format Adherence: Proper paragraph breaks, lists, or code blocks
    \end{itemize}
    
    \item \textbf{Novelty Verification} ($\mathcal{N}$)
    \begin{itemize}[nosep]
        \item Similarity Threshold: $\forall j \in \mathcal{D}, \text{sim}(i,j) \leq 0.7$
        \item Lexical Diversity: Minimum Type-Token Ratio of 0.4
        \item Response Uniqueness: No duplicate responses within same category
        \item Task Format Variation: Ensure uniform distribution across formats
    \end{itemize}
\end{enumerate}

A pair $(i,r)$ is accepted if and only if:
\[ \mathcal{F}(i,r) = \mathbb{1}[\mathcal{L}(i,r) \land \mathcal{C}(i,r) \land \mathcal{Q}(i,r) \land \mathcal{N}(i,r)] = 1 \]

This rigorous filtering ensures the quality and diversity of our final dataset while maintaining Bengali linguistic and cultural authenticity.

\clearpage

\section{Experimentation Details}
\label{app:params}




\subsection{Pretraining HyperParameters}

\begin{table}[!h]
\centering
\begin{tabular}{l c}
\toprule
\textbf{Hyperparameter} & \textbf{Value} \\
\midrule
Per device train batch size & 64 \\
Gradient accumulation steps & 16 \\
Number of training epochs & 4 \\
Learning rate & $5 \times 10^{-6}$ \\
FP16 & False \\
BF16 & True \\
Dataloader num workers & 8 \\
Gradient checkpointing & True \\
Logging steps & 1000 \\
DDP find unused parameters & False \\
Max gradient norm & 1.0 \\
Warmup steps & 1000 \\
Evaluation strategy & steps \\
Evaluation steps & 1,000 \\
Save strategy & steps \\
Save steps & 1,000 \\
Save total limit & 3 \\
Load best model at end & True \\
Metric for best model & loss \\
Greater is better & False \\
\bottomrule
\end{tabular}
\caption{Final set of hyperparameters, chosen empirically after several iterations of trial and error, for pretraining on the \texttt{Bangla-TextBook} corpus.}
\label{tab:hyperparameters}
\end{table}

\subsection{Finetuning Hyperparameters}

\begin{table}[!h]
\centering
\label{tab:hyperparameters}
\begin{tabular}{@{}ll@{}}
\toprule
Parameter & Value \\
\midrule
Max Sequence Length & 2048 \\
Batch Size (Train/Eval) & 16 \\
Gradient Accumulation Steps & 4 \\
Number of Epochs & 3 \\
Learning Rate & 1e-5 \\
Weight Decay & 0.02 \\
Warmup Steps & 10\% \\
Optimizer & AdamW (8-bit) \\
LR Scheduler & Cosine \\
Precision & BF16 \\
Evaluation Strategy & Steps \\
Evaluation Steps & 50 \\
Save Strategy & Steps \\
Save Steps & Varies \\
Seed & 42 \\
\bottomrule
\end{tabular}
\caption{Final set of hyperparameters, chosen empirically after several iterations of trial and error, for finetuning \underline{TigerLLM} (1B).}
\label{tab:hyperparameters2}
\end{table}

\begin{table}[!h]
\centering
\label{tab:hyperparameters}
\begin{tabular}{@{}ll@{}}
\toprule
Parameter & Value \\
\midrule
Max Sequence Length & 2048 \\
Batch Size (Train/Eval) & 32 \\
Gradient Accumulation Steps & 8 \\
Number of Epochs & 3 \\
Learning Rate & 1e-6 \\
Weight Decay & 0.04 \\
Warmup Steps & 15\% \\
Optimizer & AdamW (8-bit) \\
LR Scheduler & Cosine \\
Precision & BF16 \\
Evaluation Strategy & Steps \\
Evaluation Steps & 250 \\
Save Strategy & Steps \\
Save Steps & Varies \\
Seed & 42 \\
\bottomrule
\end{tabular}
\caption{Final set of hyperparameters, chosen empirically after several iterations of trial and error, for finetuning \underline{TigerLLM} (9B).}
\label{tab:hyperparameters2}
\end{table}

\end{document}